%% file: root.tex
\documentclass[letterpaper, 10 pt, conference]{ieeeconf}  % Comment this line out if you need a4paper

\IEEEoverridecommandlockouts                              % This command is only needed if 
\usepackage{subcaption}
\usepackage{graphicx}
\usepackage{comment}

\usepackage[normalem]{ulem} % for sout
\usepackage{amsmath}

\usepackage{enumerate}
\usepackage{xcolor}
\usepackage{graphicx}
\usepackage{hyperref}
\usepackage{cite}
\usepackage{adjustbox}
\usepackage{soul}

\title{\LARGE \bf
Learning Object Localization and 6D Pose Estimation from Simulation and Weakly Labeled Real Images
}

 \author{Jean-Philippe Mercier$^{1}$, Chaitanya Mitash$^{2}$, Philippe Gigu\`ere$^{1}$ and Abdeslam Boularias$^{2}$%
 \thanks{$^{1}$ Laval University, Quebec, Canada.}
 \thanks{$\enspace $ jean-philippe.mercier.2@ulaval.ca, philippe.giguere@ift.ulaval.ca}
 \thanks{$^{2}$ Rutgers University, NJ, USA}%
 \thanks{$\enspace$ \{cm1074,ab1544\}@rutgers.edu.}
 }

\begin{document}
  
%\IEEEpeerreviewmaketitle
\maketitle
\thispagestyle{empty}
\pagestyle{empty}

%%%%%%%%%%%%%%%%%%%%%%%%%%%%%%%%%%%%%%%%%%%%%%%%%%%%%%%%%%%%%%%%%%%%%%%%%%%%%%%%
\begin{abstract}
Accurate pose estimation is often a requirement for robust robotic grasping and manipulation of objects placed in cluttered, tight environments, such as a shelf with multiple objects. When deep learning approaches are employed to perform this task, they typically require a large amount of training data. However, obtaining precise 6 degrees of freedom for ground-truth can be prohibitively expensive. This work therefore proposes an architecture and a training process to solve this issue. More precisely, we present a weak object detector that enables localizing objects and estimating their 6D poses in cluttered and occluded scenes. To minimize the human labor required for annotations, the proposed detector is trained with a combination of synthetic and a few weakly annotated real images (as little as 10 images per object), for which a human provides only a list of objects present in each image (no time-consuming annotations, such as bounding boxes, segmentation masks and object poses). To close the gap between real and synthetic images, we use multiple domain classifiers trained adversarially. During the inference phase, the resulting class-specific heatmaps of the weak detector are used to guide the search of 6D poses of objects. Our proposed approach is evaluated on several publicly available datasets for pose estimation. We also evaluated our model on classification and localization in unsupervised and semi-supervised settings. The results clearly indicate that this approach could provide an efficient way toward fully automating the training process of computer vision models used in robotics.

\end{abstract}

%%%%%%%%%%%%%%%%%%%%%%%%%%%%%%%%%%%%%%%%%%%%%%%%%%%%%%%%%%%%%%%%%%%%%%%%%%%%%%%%

\input{1-intro}

\input{2-related_work}

\input{3-proposed_approach}
\input{4-experiments}
\input{5-conclusion}
\bibliography{IEEEabrv,references}
\bibliographystyle{IEEEtran}

\end{document}

%% file: 1-intro.tex
\section{Introduction}
\label{sec:intro}

\begin{figure}[t!]
      \begin{center}
      \includegraphics[width=1\linewidth]{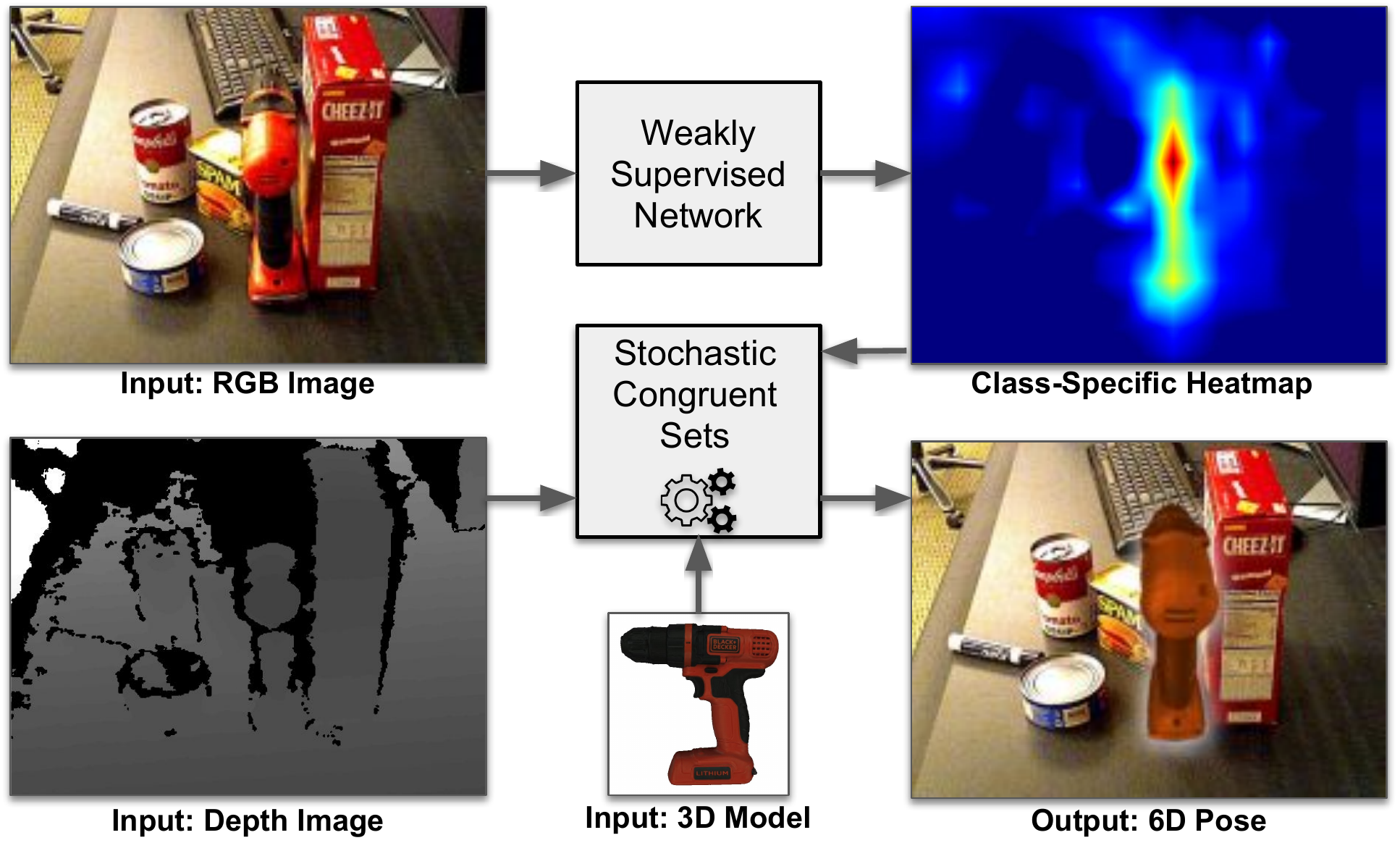}
      \end{center}
      \caption{Overview of our approach for 6D pose estimation at inference time. This figure shows the pipeline for the drill object of the YCB-video dataset~\cite{xiang2017posecnn}. A deep learning model is trained with {\it weakly annotated} images. Extracted class-specific heatmaps, along with 3D models and the depth image, guide the Stochastic Congruent Sets (StoCS) method~\cite{mitash2018robust} to estimate 6D object poses. Further details of the network are available in Section \ref{sec:approach}.}
      \label{fig_pose_estimation}
      \vspace{-0.45cm}
\end{figure}

Robotic manipulators are increasingly deployed in challenging  situations that include significant occlusion and clutter. Prime examples are warehouse automation and logistics, where such manipulators are tasked with picking up specific items from dense piles of a large variety of objects, as illustrated in Fig.~\ref{fig_pose_estimation}. The difficult nature of this task was highlighted during the recent Amazon Robotics Challenges~\cite{Correll:2016aa}. These robotic manipulation systems are generally endowed with a perception pipeline that starts with object recognition, followed by the object's six degrees-of-freedom (6D) pose estimation. It is known to to be a computationally challenging problem, largely due to the combinatorial nature of the corresponding global search problem. A typical strategy for pose estimation methods~\cite{hinterstoisser2016going,krull2015learning,brachmann2014learning,michel2017global} consists in generating a large number of candidate 6D poses for each object in the scene and refining hypotheses with the Iterative Closest Point (ICP)~\cite{besl1992method} method or its variants. The computational efficiency of this search problem is directly affected by the number of pose hypotheses. Reducing the number of candidate poses is thus an essential step towards real-time grasping of objects.

%An accurate point-wise object detector is necessary for reducing the number of candidate poses, by focusing the search only on relevant parts of the image. It thus constitute an essential step towards real-time grasping of objects. 

%\jp{I'm really not sure about the rest of this paragraph. Do we have examples of papers that do this approach?? I think the message we want to pass is that feature-based methods requires ICP (long) and that end-to-end deep learning requires time-consuming annotations.}
%(~\cite{xiang2017posecnn,hinterstoisser2016going,li2018deepim,oberweger2018making,krull2015learning,brachmann2014learning})

%As robot are increasingly deployed in uncontrolled environments such as large warehouses, the need for robustness to changes in illumination, occlusion

%This component's robustness is becoming increasingly important, as robots are being deployed in environments less structured than traditional manufacturing setups. Warehouse automation and logistics are prime examples of application domains where a robotic manipulator is tasked with picking up specific items from a dense pile of a large variety of objects, placed in complex arrangements where only a small part of each object is potentially visible to the robot's camera, as illustrated in Figure~\ref{}. The challenging nature of this task was highlighted during the recent Amazon Robotics Challenges~\cite{Correll:2016aa}. 

Training Convolutional Neural Networks (CNN) for tasks such as object detection and segmentation~\cite{Princeton,hernandez2016team,shelhamer2016fully} makes it possible to narrow down the regions that are used for searching for object poses in RGB-D images. However, CNNs typically require large amounts of annotated images to achieve a good performance. While such large datasets are publicly available for general-purpose computer vision, specialized datasets in certain areas such as robotics and medical image analysis tend to be significantly scarcer and time-consuming to obtain. In a warehouse context (our target context), new items are routinely added to inventories. It is thus impractical to collect and manually annotate a new dataset every time an inventory gets updated, particularly if it must cover all possible lighting and arrangement conditions that a robot may encounter during deployment. This is even more challenging if one wants this dataset to be collected by non-expert workers. The main goal of our approach is thus to reduce such a need for manual labeling, including completely eliminating bounding boxes, segmentation masks and 6D ground truth manual annotations.

Our first solution to reduce manual annotations is to leverage synthetic images generated with a CAD model rendered on diverse backgrounds. However, the visual features difference between real and synthetic images can be large to the point of leading to poor performance on real objects. The problem of learning from data sampled from non-identical distributions is known as {\it domain adaptation}. Domain adaptation has been increasingly seen as a solution to bridge the gap between domains~\cite{wang2018deep,csurka2017domain}. Roughly speaking, domain adaptation tries to generalize the learning from a \textit{source domain} to a \textit{target domain}, or in our case, from synthetic to real images. Since labeled data in the target domain is unavailable or limited, the standard way %of doing domain adaptation
is to train on labeled source data, while trying to minimize the distribution discrepancy between source and target domains. %%%%%%%%%%%%%%% I REMOVED THIS FROM THE ICRA SUBMISSION%%%%%%%%%%%%%%%%%%%%%%%%%
% \jp{ \textbf{TO REMOVE} One such successful approach is called Domain-Adversarial training of Neural Networks (DANN)~\cite{ganin2016domain}. 
%However, it has a detrimental tendency to align the whole feature distribution together, instead of in a class-specific way. %This in turns decreases the discriminative power of the approach. To reduce feature misalignments between classes, 
%The Multi-Adversarial Domain Adaptation (MADA)~\cite{pei2018multi} alleviates this issue by using one domain discriminator for each class, and weighting their input features by their associated class probability. % outputted by a classification module.
%Another approach, called \textit{domain randomization}, circumvents the whole issue by generating so many variant of domains that the true one can be perceived as yet another instance of domain. It has been used in the context of object detection by \cite{tremblay2018training}, in which they show that deep networks can perform as well or even better on unrealistic synthetic data than on realistic. They also show the power of fine-tuning with labeled real images, even on a small set, which we also explore in our work.}
%%%%%%%%%%%%%%% I REMOVED THIS FROM THE ICRA SUBMISSION%%%%%%%%%%%%%%%%%%%%%%%%%

While having a small labeled dataset on a target domain allows to boost performances, it may still require significant human effort for the annotations. Our second solution is to use {\it weakly supervised learning}, which significantly decreases annotation efforts, albeit with a reduced performance compared to fully-annotated images. Some methods~\cite{oquab2015object,durand2017wildcat} have been shown to be able to retrieve a high level representation of the input data (such as object localization) while only being trained for object classification. To the best of our knowledge, this promising kind of approach has not yet been applied within a robotic manipulation context.

In this paper, we propose a two-step approach for 6D pose estimation, as shown in Fig.~\ref{fig_pose_estimation}. First, we train a network for classification through domain adaptation, by using a combination of weakly labeled synthetic and real color images. During the inference phase, the weakly supervised network generates class-specific heatmaps that are subsequently refined with an independent 6D pose estimation method called Stochastic Congruent Sets (StoCS)~\cite{mitash2018robust}. Our complete method achieves competitive results on the YCB-video object dataset~\cite{xiang2017posecnn} and Occluded Linemod~\cite{krull2015learning} while using only synthetic images and few weakly labeled real images (as little as 10) per object in training. %\phil{I feel that something like Contrary to other approaches, we need significantly less data or something like that.}.
We also empirically demonstrate that for our test case, using domain adaptation in semi-supervised settings is preferable than training in unsupervised settings and fine-tuning on available weakly labeled real images, a commonly-accepted strategy when only a few images from the target domain are available.

%% file: 2-related_work.tex
\section{Related Works}
\label{sec:rel_work}

In this paper, we aim at performing object localization and 6D pose estimation with a deep network, with minimal human labeling efforts. Our approach is based on training from synthetic and weakly labeled real images, via domain adaptation. These various concepts are discussed below.

%jp There is no pre-training in the final solution (training all at once)

%TODO
% Add normalization in figure
\begin{figure*}[thpb]
      \begin{center}
      \includegraphics[width=0.75\linewidth]{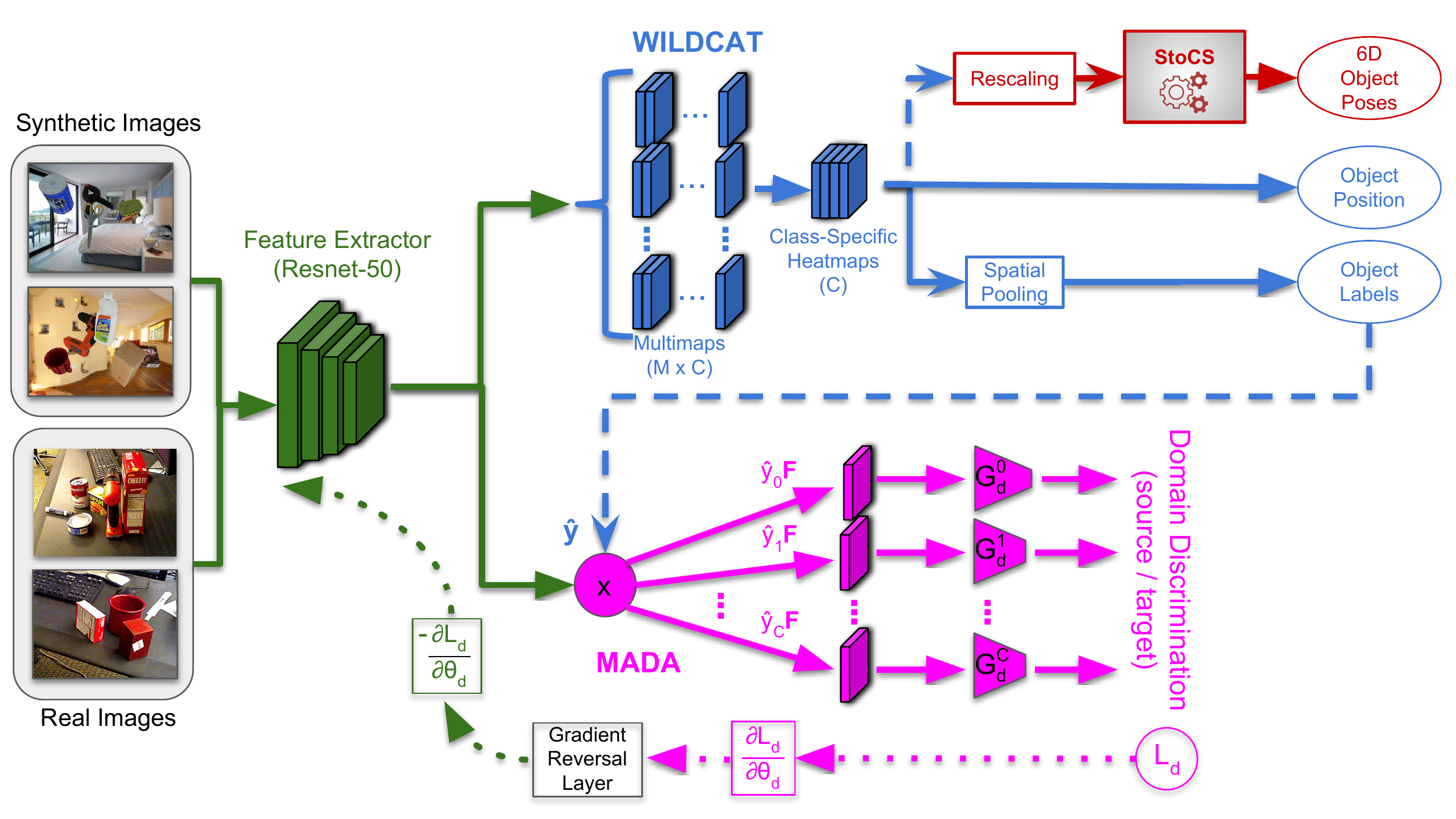}
      \end{center}
      \caption{Overview of the proposed approach for object localization and 6D pose estimation with domain adaptation, using a mix of synthetic images and weakly labeled real images.}
 \label{fig_wildcat_mada}
 \vspace{-0.55cm}
\end{figure*}
% \phil{Il manque de colle logique entre les diffÃ©rents Ã©lÃ©ments. Il faudrait expliquer le lien entre les deux approches mentionnÃ©es ici-bas et le domain adaptation ou transfert learning.}

%\phil{I changed the order of the paragraph from the original CoRL submission.}
\textbf{6D Pose Estimation} Recent literature in pose estimation focuses on learning to predict 6D poses using deep learning techniques. For example, \cite{xiang2017posecnn} predicts separately the object center in images for translation and regresses over the quaternion representation for predicting the rotation. Another approach is to first predict 3D object coordinates, followed by a RANSAC-based scheme to predict the object's pose\cite{brachmann2014learning,michel2017global}. Similarly, \cite{michel2017global} uses geometric consistency to refine the predictions from the learned model. These methods, however, need access to several images that are manually labeled with the full object poses, which is time-consuming to acquire. Some other approaches make use of the object segmentation output to guide a global search process for estimating object poses in the scene \cite{mitash2018robust, narayanan2016discriminatively, mitash2017improving}. Although the search process could compensate for errors in prediction when the segmentation module is trained with synthetic data, the domain gap could be large, and a computationally expensive search process may be needed to bridge this gap.

\textbf{Learning with Synthetic Data}
Training with synthetic data has recently gained significant traction, as shown by the multiple synthetic datasets recently available~\cite{gaidon2016virtual,mayer2016large,qiu2016unrealcv,ros2016synthia,johnson2017driving, richter2017playing}, with some focusing on optimizing the realism of the generated images. While the latter can decrease to a certain degree the gap between real and synthetic images, it somehow defeats the purpose of using simulation as a cost-effective way to create training data. To circumvent this issue, \cite{dwibedi2017cut,georgakis2017synthesizing} proposed instead to create images using segmented object instances copied on real images. This type of approach, akin to data augmentation, is however limited to the number of object views and illuminations that are available in the original dataset.
Recently,~\cite{hinterstoisser2017pre,tremblay2018training} showed promising results by training object detectors with 3D models rendered in simulation with randomized parameters, such as lighting, number of objects, object poses, and backgrounds. While in~\cite{hinterstoisser2017pre} they only uses synthetic images in training, \cite{tremblay2018training} demonstrated the benefits of fine-tuning on a limited set of real labeled images. The last one also showed that using photorealistic synthetic images does not necessarily improve object detection, compared to training on a less realistic synthetic dataset generated with randomized parameters.

%While some datasets try to mimic a corresponding real dataset, an approach called domain randomization \cite{sadeghi2016rl, tobin2017domain} has shown that image realism is not always necessary to generalize well to real images. \phil{I am not sure you should say the following, unless you make explicit link to your work:}In fact, generating images under varying conditions might prove to be better than realistic images.

%A recent approach is called domain randomization \cite{sadeghi2016rl, tobin2017domain}. For this method, many simulation parameters are randomized to generate synthetic data. The core of the idea is to generate large variations so that real data is seen as a variation of the simulation world. In \cite{tremblay2018training}, they used that idea to train an object detector on their randomized synthetic data. In their experiments, models train on their data performs better on the real dataset than when trained on \phil{check spelling:} Virual Kitty, which tries to mimic the real dataset. It therefore shows that realism is not necessarily important to always bridge the reality gap. They also investigated the idea of \cite{hinterstoisser2017pre}, in which they claim that freezing the weights of the early layers of a pre-trained network is helpful. However, in their experiments, they found it to be more harmful than helpful. They also reported that a huge improvement by using pretrained networks instead of training from scratch, even with huge datasets of 1 million images. Their performance saturates with synthetic dataset size of 10k images when pretrained. 

\textbf{Domain Adaptation}
Domain adaptation techniques~\cite{wang2018deep,csurka2017domain} can serve to decrease the distribution discrepancy between different domains, such as real vs. synthetic. The popular DANN~\cite{ganin2016domain} approach relies on two classifiers: one for the desired task, trained on labeled data from a source domain, and another one (called \emph{domain classifier}) that classifies whether the input data is from the source or target domain. Both classifiers share the first part of the network, which acts as a feature extractor. The network is trained in an adversarial manner: domain classifier parameters are optimized to minimize the domain classification loss, and shared parameters are optimized to maximize the domain classification loss. It is possible to achieve this minimax optimization in a single step by using a gradient reversal layer that reverses the sign of the gradient between shared and non-shared parameters of the domain classifier. To the best of our knowledge, the present work is the first use a DANN-like approach for point-wise object localization, a fundamental problem in robotic manipulation. 

% DANNs \cite{ganin2016domain} are one of the methods commonly used to make features independent between an input domain and a target domain. The network can be divided into 3 parts: feature extractor, task network and domain classifier. The goal of the architecture is to be able to have good classification on the source domain, while being unable to differentiate between domains. The latter is achieved by using a gradient reversal layer between the feature extractor and the domain classifier. In semi-supervised scenarios, the task network is trained on the annotated source data (annotated target data not available).

% \jp{Domain Randomization: C'est pratiquement ce qu'on fait. On randomise les backgrounds, les objets, leur position/rotation. Durant le training, on fait de l'augmentation qui randomise le scale, couleur/contraste/eclairage. Je pense que c'est pour cette raison que la detection fonctionne aussi bien sur les donnes synthetiques seulement}

% In \cite{rad2017feature}, they observed that for paired images, only a few feature coefficients were \phil{je comprends pas: creating the domain gap}. To solve the problem, they trained a network to minimize the $L_2$ loss between synthetic image features and transformed real ones. In their case, it reduced the large differences that caused the domain gap, thus allowing them to train their network for pose estimation. They have also shown that their method outperforms DANN for pose estimation on LineMod dataset.

\textbf{Weakly Supervised Learning}
We are interested in weakly supervised learning with inexact supervision, for which only coarse-grained labels are available~\cite{zhou2017brief}. In~\cite{oquab2015object}, a network was trained only with weak image-level labels (classes that are present in images, but not their position) and max-pooling was used to retrieve approximate location of objects. The proposed {\it WILDCAT} model~\cite{durand2017wildcat} performs classification and weakly supervised point-wise detection and segmentation. This architecture learns multiple localized features for each class, and uses a spatial pooling strategy that generalizes to many ones (max pooling, global average pooling and negative evidence). In the present work, we push the paradigm of minimum human supervision \emph{even further}. To this effect, we propose to train WILDCAT with synthetic images, in addition to weakly supervised real ones, and use MADA (a variant of DANN) for domain adaptation.

%% file: 3-proposed_approach.tex
\section{Proposed Approach}
\label{sec:approach}
We present here our approach to object localization and 6D pose estimation. It is trained using a mix of synthetic and real images and only requires weak annotations (only class-presence) in both domains. 

\vspace{-0.55cm}
\subsection{Overview}

% \begin{figure}[thpb]
%       \begin{center}
%       \fbox{\includegraphics[width=0.98\linewidth]{images/pose_estimation_pipeline_cropped}}
%       \end{center}
%       \caption{Pose estimation pipeline}}
%  \label{fig_pose_estimation}
% \end{figure}

Figure~\ref{fig_wildcat_mada} depicts an overview of our proposed system. It comprises {\it i}) a {\it ResNet-50} model pre-trained on {\it ImageNet} as a feature extractor (green), {\it ii}) a weak classifier inspired from the WILDCAT model~\cite{durand2017wildcat} (blue), {\it iii}) the Stochastic Congruent Sets ({\it StoCS}) for 6D pose estimation (red)~\cite{mitash2018robust}, and {\it iv}) the MADA domain adaptation network to bridge the gap between synthetic and real data. %\sout{derived from Multi-Adversarial Domain Adaptation network ({\it MADA})~\cite{pei2018multi} with a gradient reversal layer to learn domain-independent features. } \sout{The different parts of the system and the training process are explained in the following sections.}
%The ResNet extracts features from the images that are useful for object classification (\phil{, localization and pose estimation}. \sout{which could also be used for point-wise localization by WILDCAT.} 
During the inference phase, the domain adaptation part of the network is discarded. Given a test image, class-specific heatmaps are generated by the network. These heatmaps indicate the most probable locations of each object in the image. This probability distribution is then fed to StoCS, a robust pose estimation algorithm that is specifically designed to deal with noisy localization.
%\phil{Here i feel you re-explain what DANN is, which you have done before. Focus should be on the difference with MADA.} 
To force the feature extractor to extract similar features for both synthetic and real images, a MADA module (described below) is employed.
%is trained to classify any input image according to its domain $\{\mathrm{real, synthetic} \}$ and the weights of ResNet are updated such that this classification's error increases. 
MADA's purpose is to generate gradients during training (via a reversal layer) in order to improve the generalization capabilities of the feature extractor. 

%\phil{CONFIRM THE FOLLOWING!!! Once training is achieved, the MADA module is discarded.} ---> I've moved it higher in the paragraph (JP)

% Thus, ResNet is trained to extract features that make it hard for MADA to distinguish between synthetic and real images. \phil{This is not part of architecture, but training regimen: MADA's role ends once the training of ResNet and WILDCAT is over.} 
%\phil{Are all parts active at all times? Are some only used during training? during testing?}

% TODO
% Add architecture and training regimen in another paragraph

\subsection{Synthetic Data Generation}
For synthetic data generation, we used a modified version of the SIXD toolkit\footnote{\url{https://github.com/thodan/sixd_toolkit}}. This toolkit generates color and depth images of 3D object models rendered on black backgrounds. Virtual camera viewpoints are sampled on spheres of different radii, following the approach described in~\cite{hinterstoisser2008simultaneous}. We extended the toolkit with the functionality of rendering more than one object per image, and also used random backgrounds taken from the LSUN dataset~\cite{YuZSSX15}. % instead of generic black backgrounds.
Similarly to recent {\it domain randomization} techniques~\cite{DBLP:conf/iros/TobinFRSZA17}, we observed from our experiments that these simple modifications help transferring from simulation to real environments where there are multiple objects of interest, occlusions and diverse backgrounds. Figure~\ref{fig_wildcat_mada} displays some examples of the generated synthetic images that we used to train our network.

% \begin{figure}
%  \begin{center}
%   \begin{subfigure}[b]{0.25\textwidth}
%     \includegraphics[width=\textwidth]{images/syn_ycb_1}
%   \end{subfigure}
%   % 
%   \begin{subfigure}[b]{0.25\textwidth}
%     \includegraphics[width=\textwidth]{images/syn_ycb_2}
%   \end{subfigure}
%   %
%   \begin{subfigure}[b]{0.25\textwidth}
%     \includegraphics[width=\textwidth]{images/syn_ycb_3}
%   \end{subfigure}
% \end{center}
%   \caption{Examples of synthetic images used to train the proposed system for the {\it YCB} objects~\cite{calli2017yale}.}
%   \label{fig_synthetic_images}
% \end{figure}

\subsection{Weakly Supervised Learning with WILDCAT}
The images used for training our system are weakly labeled: only a list of object classes present in the image is provided. In order to recover localization from such weak labels, we leverage the WILDCAT architecture~\cite{durand2017wildcat}. Indeed, WILDCAT is able to recover localization information through its high-level feature map, even though it is only trained with a classification loss. %\sout{Despite being only trained for classification, WILDCAT implicitly recovers localization information for each object through  responses in specific feature maps. }%, which is illustrated in blue in Figure \ref{fig_wildcat_mada}. 
As a feature extractor, we employ a ResNet-50 (pretrained on ImageNet) for which the last layers (global average pooling and fully connected layers) are removed, as depicted in Figure~\ref{fig_wildcat_mada}. The WILDCAT architecture added on top of this ResNet-50 comprises three main modules: a \textit{multimap transfer layer}, a \textit{class pooling layer} and a \textit{spatial pooling layer}. The \textit{multimap transfer layer} consists of $1 \times 1$ convolutions that extracts $M$ class-specific modalities per class $C$, with $M=8$ as per the original paper~\cite{durand2017wildcat}. The \textit{class pooling} module is an average pooling layer that reduces the number of feature maps from $MC$ to $C$. Then, the \textit{spatial pooling} module selects $k$ regions with maximum/minimum activations to calculate scores for each class. The classification loss for this module is a multi-label one-versus-all loss based on max-entropy (\textit{MultiLabelSoftMarginLoss} in PyTorch). The classification scores are then rescaled between 0 and 1 to cooperate with MADA.

\subsection{Multi-Adversarial Domain Adaptation with MADA}
We used the {\it Multi-Adversarial Domain Adaptation} (MADA) approach~\cite{pei2018multi} to bridge the ``reality gap''. MADA extends the {\it Domain Adversarial Networks} (DANN) approach~\cite{ganin2016domain} by using one domain discriminator per class, instead of a single global discriminator as in the original version of DANN~\cite{ganin2016domain}. Having one discriminator per class has been found to help aligning class-specific features between domains. In MADA, the loss $L_d$ for the $K$ domain discriminators and input $\mathbf{x_i}$ is defined as:
\begin{equation}
L_d = \frac{1}{n} \sum_{k=1}^{K} \sum_{\mathbf{x_i} \in D_s \cup D_t} L_d^k\bigg(G_d^k \Big(\hat{y}_i^k G_f(\mathbf{x_i})\Big), d_i\bigg),
\label{eq_mada_loss}
\end{equation}
wherein $i \in \{1, \dots, n\}$, and $n = n_s + n_t$ is the total number of training images in source domain $D_s$ (synthetic images) and the target domain $D_t$ (real images). $G_f$ is the feature extractor (the same for both domains), $\hat{y}_i^k$ is the probability of label $k$ for image $\mathbf{x_i}$. This probability $\hat{y}_i^k$ is the output of the weak classifier WILDCAT. $G_d^k$ is the $k$-th domain discriminator and $L_d^k$ is its cross-entropy loss, given the ground truth domain $d_i\in \{\mathrm{synthetic, real}\}$ of image $\mathbf{x_i}$. Our global objective function is:
% \begin{equation}
% C = \frac{1}{n_s} \sum_{\mathbf{x_i} \in D_s } L_y \bigg( G_y \Big( G_f(\mathbf{x_i}) \Big), y_i \bigg) - \frac{\lambda}{n} \sum_{k=1}^{K} \sum_{\mathbf{x_i} \in D} L_d^k\bigg(G_d^k \Big(\hat{y}_i^k G_f(\mathbf{x_i})\Big), d_i\bigg).
% \label{eq_mada_loss_global}
% \end{equation}
%\begin{multline}
\begin{equation}
C = \frac{1}{n} \sum_{\mathbf{x_i} \in D } L_y \bigg( G_y \Big( G_f(\mathbf{x_i}) \Big), y_i \bigg) - \lambda L_d  \quad , 
%  - \frac{\lambda}{n} \sum_{k=1}^{K} \sum_{\mathbf{x_i} \in D} L_d^k\bigg(G_d^k \Big(\hat{y}_i^k G_f(\mathbf{x_i})\Big), d_i\bigg)
\label{eq_mada_loss_global}
\end{equation}
%\end{multline}
where $L_y$ is the classification loss, $L_d$ the domain loss and $\lambda$ has been found to work well with a value of 0.5. The heat-map probability distribution extracted from WILDCAT is used to guide the StoCS algorithm in its search for 6D poses, as explained in the next section.  %\phil{I have simplified above equation. Please confirm! How do you find the value of $\lambda$?} (\jp{pretty much trial-and-error, so many hyperparameters...})

\subsection{Pose Estimation with Stochastic Congruent Sets (StoCS)}
The StoCS method~\cite{mitash2018robust} is a robust pose estimator that predicts the 6D pose of an object in a depth image from its 3D model and a probability heatmap. 
%\phil{In order to employ the score heatmap generated from WILDCAT, we developed a specific normalization approach (as earlier attempts at using unnormalized heatmaps were unsuccessful.)} 
We employ a min-max normalization on the class-specific heatmaps of the Wildcat network, transforming them into a probability heatmaps $w_{p_i}$, using the per-class minimum ($w_{min}$) and maximum ($w_{max}$) values:
%it is generated by normalizing over an intermediate output of the WILDCAT network with activation $w_{p_i}$, 
\begin{equation}
\label{eq_normalization}
\pi_{p_i \rightarrow O_k} = \frac{w_{p_i} - w_{min}}{w_{max} - w_{min}}.
\end{equation}
This generates a heatmap providing the probability $\pi$ of an object $O_k$ being located at a given pixel $p_i$. The StoCS algorithm then follows the paradigm of a randomized alignment technique. It does so by iteratively sampling a set of four points, called a base $B$, on the point cloud $S$ and finds corresponding set of points on the object model $M$. Each corresponding set of four points defines a rigid transformation $T$, for which an alignment score is computed between the transformed model cloud and the heatmap for that object. The optimization criteria is defined as\\
\begin{eqnarray}
T_{opt} = arg\,max_{T}\sum_{m_i \in M_k}f( m_i, T,
S_k),\\
f(m_i, T, S_k ) =
\pi_k(s*), if \mid T(m_i) - s* \mid < \delta_s.
\end{eqnarray}
The base sampling process in this algorithm considers the joint probability of all four points belonging to the object in question, given as\\
\begin{equation}
Pr(B \rightarrow O_k) = \frac{1}{Z}
\prod_{i=1}^{4} \{ \phi_{node}(b_i) \prod_{j=1}^{j<i} \phi_{edge}(b_i,
b_j) \}.
\end{equation}
where $\phi_{node}$ is obtained from the probability heatmap and $\phi_{edge}$ is computed based on the point-pair features of the pre-processed object model. Thus, the method combines the normalized output of the Wildcat network with the geometric model of objects to obtain base samples which belong to the object with high probability.

In the next two Sections, we demonstrate the usefulness of our approach. First in Section~\ref{sec:w_exp}, we quantify the importance of each component (Wildcat, MADA) in order to train a network that generates \emph{relevant} feature maps from weakly labeled images. In Section~\ref{sec:6d_exp}, we then evaluate the performance of using these heatmaps with StoCS for rapid 6D pose estimation, which is the final goal of our paper.

% \begin{figure}[thpb]
%       \begin{center}
%       \fbox{\includegraphics[width=0.4\linewidth]{images/Stocs_Overview}}
%       \end{center}
%       \caption{Input (left) and output (right) of Stochastic Congruent Sets (StoCS) for 6D pose estimation}
%       \label{fig_stocs_overview}
% \end{figure}

%WSL transfer layer (see Fig. \ref{fig_wildcat}). 

% \phil{Justifier pourquoi avoir pris un ResNet-50 au lieu d'autres features extractors (ResNet-xxx, DenseNet, etc).}

%% file: 4-experiments.tex
\section{Weakly Supervised Learning Experiments for object detection and classification}
\label{sec:w_exp}

% \begin{figure}
%  \begin{center}
%   \begin{subfigure}[b]{0.2\textwidth}
%     \includegraphics[width=\textwidth]{images/img01_rgb}
%     \caption{Weak/Class labels}
%     \label{fig_supervision_rgb}
%   \end{subfigure}
%   % 
%   \begin{subfigure}[b]{0.2\textwidth}
%     \includegraphics[width=\textwidth]{images/img01_bbox}
%     \caption{Bounding boxes}
%     \label{fig_supervision_bbox}
%   \end{subfigure}
%   %
%   \begin{subfigure}[b]{0.2\textwidth}
%     \includegraphics[width=\textwidth]{images/img01_seg}
%     \caption{Pixelwise}
%     \label{fig_supervision_seg}
%   \end{subfigure}
%   %
%   \begin{subfigure}[b]{0.2\textwidth}
%     \includegraphics[width=\textwidth]{images/img01_pose}
%     \caption{6D poses}
%     \label{fig_supervision_pose}
%   \end{subfigure}
%  \end{center}
% \caption{Different supervision levels for training pose estimation methods. In (a), class of objects in the image are known. }
% \label{fig_supervision_levels}
% \end{figure}

In this first experimental section, we perform an ablation study to evaluate the impact of various components for classification and point-wise localization. We first tested our approach %with unsupervised learning (no human annotations at all),
without any human labeling, as a baseline. We then evaluated the gain obtained by employing various numbers of weakly labeled images for four semi-supervised strategies. % \sout{when an increasing number of weakly labeled real images are available. These experiments serve to demonstrate that minimalistic human effort can results in significant gains, as well as highlight the winning strategies.}

We performed these evaluations on the YCB-video dataset~\cite{xiang2017posecnn}. This dataset contains 21 objects with available 3D models. It also has full annotations for detection and pose estimation on 113,198 training images and 20,531 test images. A subset of 2,949 test images (keyframes) is also available. Our results are reported for this more challenging subset, since most images in the bigger test set are video frames that are too similar and would report optimistic results.

For these experiments, we trained our network for 20 epochs (500 iterations per epoch) with a batch size of 4 images per domain. We used stochastic gradient descent with a learning rate of $0.001$ (decay of $0.1$ at epochs 10 and 16) and a Nesterov momentum of $0.9$. The ResNet-50 was pre-trained on ImageNet and the weights of the first two blocks were frozen.

\subsection{Unsupervised Domain Adaptation}
\label{UnsupervisedDA}
\begin{figure}
 %\begin{center}
  \begin{subfigure}[b]{0.40\textwidth}
    \includegraphics[width=\textwidth]{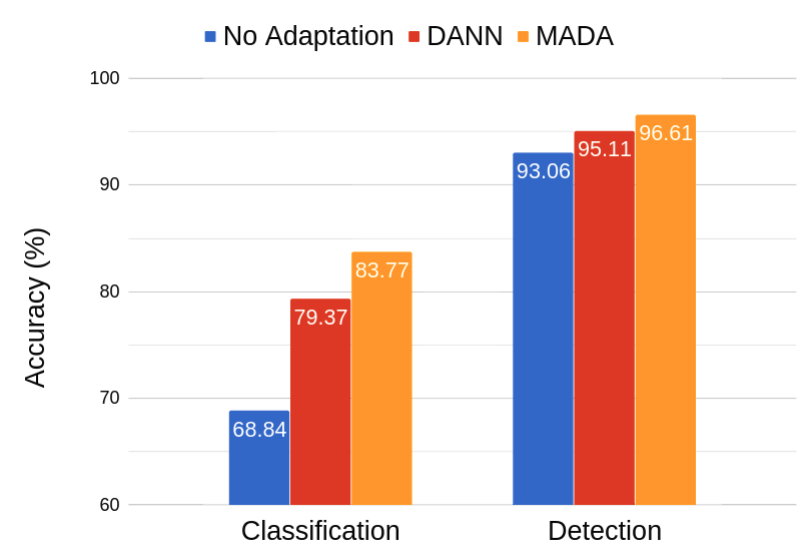}
  	\vspace{-0.5cm}
    \caption{}
      %\caption{Unsupervised domain adaptation performance for classification accuracy and point-wise detection}
      \label{fig_unsupervised_da_performance}
  \end{subfigure}
  \begin{subfigure}[b]{0.45\textwidth}
    \includegraphics[width=\textwidth]{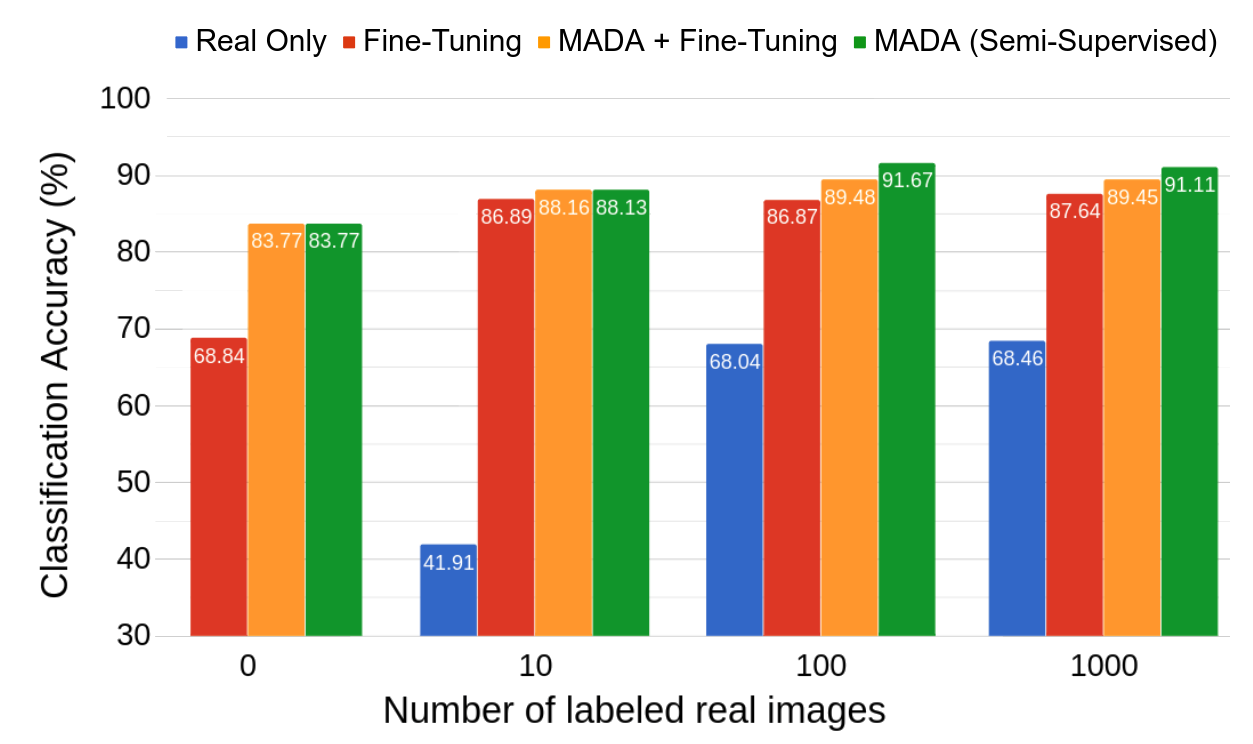}
    \caption{}
    %\caption{Classification accuracy with respect to the number of weakly labeled real images per class}
      \label{fig_classification_num_images}
  \end{subfigure}
%\end{center}
  \caption{Performance analysis. In (a), we compare classification accuracy and point-wise detection when no label on real images are available. In (b), we compare the performance of different training processes when different numbers of real images are weakly labeled. }
  \label{fig_synthetic_images}
    	\vspace{-0.55cm}
\end{figure}

For this experiment, we trained our model with weakly labeled synthetic images ($WS$) and unlabeled real images ($UR$). %\sout{from the training set of YCB-video dataset~\cite{xiang2017posecnn}}.
We tested three architecture configurations of domain adaptation: 1) without any domain adaptation module (WILDCAT model trained on $WS$), 2) with DANN ($WS$+$UR$) and 3) with MADA ($WS$+$UR$). We evaluated each of these configurations for both classification and detection. For classification, we used the accuracy metric to evaluate our model's capacity to discriminate which objects are in the image. We used a threshold of 0.5 on classification scores to predict the presence or absence of an object. For detection, we employed the point-wise localization metric~\cite{oquab2015object}, which is a standard metric to evaluate the ability of weakly supervised networks to localize objects. For each object in the image, the maximum value in their class-specific heatmap was used to retrieve the corresponding pixel in the original image. If this pixel is located inside the bounding box of the object of interest, it is counted as a good detection. Since the class-specific heatmap is a reduced scale of the input image due to pooling, a tolerance equal to the scale factor was added to the bounding box. In our case, a location in the class-specific heatmaps corresponds to a region of 32 pixels in the original image.
In Figure~\ref{fig_unsupervised_da_performance}, we report the average scores of the last 5 epochs over 3 independent random runs for each network variation. %\jp{Should we modify this??}
These results \emph{a)} confirm the importance of employing a domain adaptation strategy to bridge the reality gap, and \emph{b)} the necessity of having one domain discriminator $G_d^k$ for each of the X objects in the YCB database (MADA), instead of a single one (DANN). Next, we evaluate the gains obtained by employing weakly-annotated real images.

%TODO
% Divide the graphs into 2 separate graphs with their own descriptions

% \begin{figure}[thpb]
%       \begin{center}
%       \fbox{\includegraphics[width=0.6\linewidth]{images/Unsupervised_DA_performance}}
%       \end{center}
%       \caption{Unsupervised domain adaptation performance for classification accuracy and point-wise detection}
%       \label{fig_unsupervised_da_performance}
% \end{figure}

%TODO (add "Weakly" to "number of labeled real images" in figure b) 
\begin{table*}
\begin{center}
\resizebox{0.98\textwidth}{!}{%
\begin{tabular}{|c|c|c|c|c|c|}
\hline 
\textbf{Method} & \textbf{Modality} & \textbf{Supervision} & \textbf{Full Dataset} & \textbf{Accuracy (\%)}  &  \textbf{Accuracy (\%)} \\
 &  & &  & \textbf{YCB-Video}  & \textbf{Occluded Linemod} \\
\hline
PoseCNN \cite{xiang2017posecnn} & RGB & Pixelwise labels + 6D poses & Yes & 75.9  & 24.9 \\
\hline
PoseCNN+ICP \cite{xiang2017posecnn} & RGBD & Pixelwise labels + 6D poses & Yes & 93.0  & \textbf{78.0} \\
\hline
DeepHeatmaps \cite{oberweger2018making} & RGB & Pixelwise labels + 6D poses & Yes & 81.1 & 28.7  \\
\hline
FCN + Drost et. al. \cite{drost2010model} & RGBD & Pixelwise labels & Yes & 84.0  & - \\
\hline
FCN + StoCS \cite{mitash2018robust} & RGBD & Pixelwise labels & Yes & 90.1  & - \\
\hline
Brachmann et al. \cite{brachmann2014learning} & RGBD & Pixelwise labels + 6D poses & Yes & - & 56.6 \\ \hline
Michel et. al. \cite{michel2017global} & RGBD & Pixelwise labels + 6D poses & Yes & - & 76.7\\
\hline

\hline
OURS & RGBD & Object classes & No (10 weakly labeled images) & 88.7  &  68.8 \\
OURS & RGBD & Object classes & Yes & 90.2  & - \\
OURS (multiscale inference) & RGBD & Object classes & No (10 weakly labeled images) & -  & 76.6 \\
OURS (multiscale inference) & RGBD & Object classes & Yes & \textbf{93.6}  & - \\
\hline
\end{tabular}
}
\end{center}
\caption{Area under the accuracy-threshold curve for 6D Pose estimation on YCB-Video dataset and Occluded Linemod} \label{pose_ycb}
\vspace{-0.15cm}
\end{table*}

\vspace{-0.05cm}
\subsection{Semi-Supervised Domain Adaptation}

A significant challenge for agile deployment of robots in industrial environments is that they ideally should be trained with limited annotated data, both in terms of numbers of images and of their extensiveness of labeling (no pose information, just class). We thus evaluated the performance of four different strategies as a function of the number of such weakly-labeled real images:
%\vspace{-0.05in}
\begin{enumerate}
\setlength\itemsep{-0.02em}
\item Without domain adaptation:
	%\vspace{-0.05in}
	\begin{enumerate}
    \setlength\itemsep{-0.02em}
	\item Real Only: Trained only on weakly labeled real images,
    \item Fine-Tuning: Trained on synthetic images and then fine-tuned on weakly labeled real images,
	\end{enumerate}
\item With domain adaptation:
	%\vspace{-0.05in}
	\begin{enumerate}
    \setlength\itemsep{-0.02em}
    \setcounter{enumi}{2}
	\item Fine-Tuning: Trained on synthetic images and then fine-tuned on weakly labeled real images,
    \item Semi-Supervised: Trained with synthetic images and weakly labeled real images simultaneously.
	\end{enumerate}
   
% \item Real Only: Only real images are used in training
% \item No Adaptation + Fine-Tuning: Only the WILDCAT model is used in training. 
% \item MADA + Fine-Tuning: Our full approach is used. We first train with synthetic images and unlabeled real images. Then, we fine-tune on weakly labeled real images.
% \item MADA (Semi-Supervised): We use our full approach with labeled synthetic and real images at the same time.
\end{enumerate}

For 1.a and 1.b, we validate that using fine-tuning on a network pre-trained with synthetic data is preferable to training directly on real images. For 2.a and 2.b, we compare the performance of our approach trained with fine-tuning, and in a semi-supervised way (using images from both domains at the same time). We are particularly interested in comparing the two approaches 2.a and 2.b, since \cite{oliver2018realistic} achieved the lowest error rate compared to any other semi-supervised approach by only using fine-tuning. 

Our results are summarized in Figure~\ref{fig_classification_num_images}. From them, we conclude that training with synthetic images improves classification accuracy drastically, especially when few labels are available. Also, our approach performs slightly better when trained in a semi-supervised setting (2.b) than with a fine-tuning approach (2.a), which is contrary to~\cite{oliver2018realistic}.
%where the model is trained in unsupervised settings and fine-tuned on labeled real images .

In this Section, we justified our architecture, as well as the training technique employed, to create a network capable of performing object identification and localisation through weak learning. In the next Section, we demonstrate how the feature maps extracted by our network can be employed to perform precise 6 DoF object pose estimation via StoCS.

\section{6D Pose Estimation Experiments}
\label{sec:6d_exp}
We evaluated our full approach for 6D pose estimation on YCB-video~\cite{xiang2017posecnn} and Occluded Linemod~\cite{krull2015learning} datasets. We used the most common metrics to compare with similar methods. The average distance (ADD) metric \cite{hinterstoisser2012model} measures the average distance between the pairwise 3D model points transformed by the ground truth and predicted pose. For symmetric objects, the ADD-S metric measures the average distance using the closest point distance. Also, the visible surface discrepancy \cite{hodan2018bop} compares the distance maps of rendered models for estimated and ground-truth poses.

We used the same training details mentionned in section \ref{sec:w_exp}. Since the network architecture is fully convolutional, we also added an experiment for which we combined the output of the network for 3 different scales of the input image (at test time only).

\subsection{YCB-Video Dataset} % 2 568 images per object x 21 objects = 53 928 images
%\phil{Please specify what is different here than the dataset described in the previous section: }
This dataset comprises several frames from 92 video sequences of cluttered scenes created with 21 YCB objects. The training for competing methods \cite{xiang2017posecnn, oberweger2018making, drost2010model} is performed using 113,199 frames from 80 video sequences with semantic (pixelwise) and pose labels. For our proposed approach, we used only 10 randomly sampled weakly annotated (class labels only) real images per object class combined with synthetic images. As in \cite{xiang2017posecnn}, we report the area under the curve (AUC) of the accuracy-threshold curve, using the ADD-S metric.
%The evaluation metric proposed in the dataset benchmark \cite{xiang2017posecnn} was used to report the accuracy. It uses the average distance (ADD) metric to plot the accuracy-threshold curve, and the area under the curve (AUC) is reported. The ADD metric measures the average distance between model points transformed by the ground truth and predicted pose. 
Results are reported in Table \ref{pose_ycb}. Our proposed method achieves 88.67\% accuracy with a limited number of weakly labeled images and up to 93.60\% when using the full dataset with multiscale inference. It outperforms competing approaches, with the exception of PoseCNN+ICP, which performs similarly. However, our approach has a large computational advantage with an average runtime of 0.6 seconds per object as opposed to approximately 10 seconds per object for the modified-ICP refinement for PoseCNN. It also uses \emph{a)} nearly a hundredfold less real data, and \emph{b)} also only using the class labels. This results thus demonstrate that we can reach \emph{fast} and \emph{competitive} results without the need of 6D fully-annotated real datasets.

\subsection{Occluded Linemod Dataset} 
This dataset contains 1215 frames from a single video sequence with pose labels for 9 objects from the LINEMOD dataset with high level of occlusion. Competing methods are trained using the standard LINEMOD dataset, which consists in average of 1220 images per object. In our case, we used 10 real random images per object (manually labelled) on top of the generated synthetic images, using the weak (class) labels only. 
%In this case, we did not use the full training dataset, since Linemod only has annotations for one object per image.
As reported in Table \ref{pose_ycb}, our method achieved scores of 68.8\% and 76.6\% (multiscale) for the ADD evaluation metric and using a threshold of $10\%$ of the 3D model diameter. 
These results compare with state-of-the-art methods while using less supervision and a fraction of training data. The multiscale variant (input image at 3 different resolutions) made our approach more robust to occlusions. We did not train with the full Linemod training dataset, since the dataset only has annotations for 1 object per image and our method requires the full list of objects that are in the image. Furthermore, we evaluated our approach on the 6D pose estimation benchmark~\cite{hodan2018bop} using the visual discrepency metric. We evaluated our network with multiscale inference and we can see in Table \ref{table_vsd} that we are among the top 3 for the recall score while being the fastest. We also tested the effect of combining ICP with StoCS. At the cost of more processing time, we obtain the best performance among the methods that were evaluated on the benchmark.

% Whereas it outperforms all approaches using RGB only data, its performance is slightly inferior to some competing methods that use RGBD. We suspect the sensor noise leads to error-prone surface normal computation, which are extensively used by the StoCS approach for base sampling and in the optimization cost. 

\vspace{-0.35cm}
\begin{minipage}[t]{8cm} 
    %\begin{center}
    \centering
        \begin{adjustbox}{center, margin=0 0 0 20px}  
        \begin{tabular}{|c|c|c|}
        \hline 
        \textbf{Method} & \textbf{Recall Score (\%)} & \textbf{Time (s)}\\
        \hline
        Vidal-18 \cite{vidal20186d} & 59.3 & 4.7 \\
        Drost-10 \cite{drost2010model} & 55.4 & 2.3 \\
        Brachmann-16 \cite{brachmann2016uncertainty} & 52.0 & 4.4 \\
        Hodan-15 \cite{hodavn2015detection} & 51.4 & 13.5 \\
        Brachmann-14 \cite{brachmann2014learning} & 41.5 & 1.4 \\
        Buch-17-ppfh \cite{buch2017rotational} & 37.0 & 14.2 \\
        Kehl-16 \cite{kehl2016deep} & 33.9 & 1.8 \\
        \hline
        OURS (multiscale) & 55.2 & 0.6 \\
        OURS (multiscale) + ICP & \textbf{62.1} & 6.4 \\
        \hline
        \end{tabular}
        \end{adjustbox}
        \captionof{table}{Visual discrepency recall scores (\%) (correct pose estimation) for $\tau = 20$mm and $\theta = 0.3$ on Occluded Linemod, based on the 6D pose estimation benchmark \cite{hodan2018bop}.}
        \label{table_vsd}
        \vspace{-0.25cm}
    %\end{center}
\end{minipage}

%% file: 5-conclusion.tex
\section{Conclusion}
\label{sec:conclu}

In this paper, we explored the problem of 6D pose estimation in the context of limited annotated training datasets. To this effect, we demonstrated that the output of a weakly-trained network is sufficiently rich to perform full 6D pose estimation. Pose estimation experiments on two datasets showed that our approach is competitive with recent approaches (such as PoseCNN), despite requiring \emph{significantly less annotated images}. Most importantly, our annotation level requirement for real images is \emph{much weaker}, as we only need a class label without any spatial information (either bounding box or full 6D ground truth). %By employing the StoCS algorithm, we also circumvented the need to employ time-consuming pose estimation algorithms such as ICP. 
In this end, this makes our approach compatible with an agile automated warehouse, where new objects to be manipulated are constantly introduced in a training database by non-expert employees.